\documentclass[10pt,twocolumn,letterpaper]{article}

\usepackage{iccv}
\usepackage{times}
\usepackage{epsfig}
\usepackage{graphicx}
\usepackage{amsmath}
\usepackage{amssymb}

\usepackage{multirow}
\usepackage{enumitem}
\usepackage{siunitx}
\usepackage{comment} 
\usepackage{booktabs}
\newcommand{\RNum}[1]{\lowercase\expandafter{\romannumeral #1\relax}}

\usepackage[pagebackref=true,breaklinks=true,letterpaper=true,colorlinks,bookmarks=false]{hyperref}

\iccvfinalcopy 

\def\iccvPaperID{****} 

\ificcvfinal\fi

\begin{document}

\title{LAformer: Trajectory Prediction for Autonomous Driving with Lane-Aware Scene Constraints}

\author{Mengmeng Liu$^{1,4}$, Hao Cheng$^{2,*}$, Lin Chen$^3$, Hellward Broszio$^3$, \\
Jiangtao Li$^4$, Runjiang Zhao$^4$, Monika Sester$^1$, Michael Ying Yang$^2$  \\
$^1$Leibniz  University  Hannover, $^2$University of Twente, $^3$VISCODA GmbH, $^4$PhiGent Robotics \\
{\tt\small $^{*}$Corresponding author, h.cheng-2@utwente.nl}
}

\maketitle
\ificcvfinal\thispagestyle{empty}\fi

\begin{abstract}
   Trajectory prediction for autonomous driving must continuously reason the motion stochasticity of road agents and comply with scene constraints.
   Existing methods typically rely on one-stage trajectory prediction models, which condition future trajectories on observed trajectories combined with fused scene information. 
   However, they often struggle with complex scene constraints, such as those encountered at intersections. 
   To this end, we present a novel method, called LAformer.
   It uses a temporally dense lane-aware estimation module to select only the top highly potential lane segments in an HD map, which effectively and continuously aligns motion dynamics with scene information, reducing the representation requirements for the subsequent attention-based decoder by filtering out irrelevant lane segments.
   Additionally, unlike one-stage prediction models, LAformer utilizes predictions from the first stage as anchor trajectories and adds a second-stage motion refinement module to further explore temporal consistency across the complete time horizon.
   Extensive experiments on Argoverse 1 and nuScenes demonstrate that LAformer achieves excellent performance for multimodal trajectory prediction. 
 \end{abstract}

\section{Introduction}
\label{sec:intro}
Accurate trajectory prediction is paramount for enabling autonomous driving in diverse traffic scenarios involving interactions with various road agents. Due to the stochastic behaviors of agents and their mutual influences, in addition to the varying environmental scene contexts, trajectory prediction remains an exceedingly challenging task. Therefore, this task necessitates effective learning of an agent's motion dynamics and interactions with other agents, as well as careful consideration of scene constraints.
\begin{figure}[t]
  \centering
  \includegraphics[clip=true, trim=0.1in 2in 0.8in 0in, width=1\linewidth]{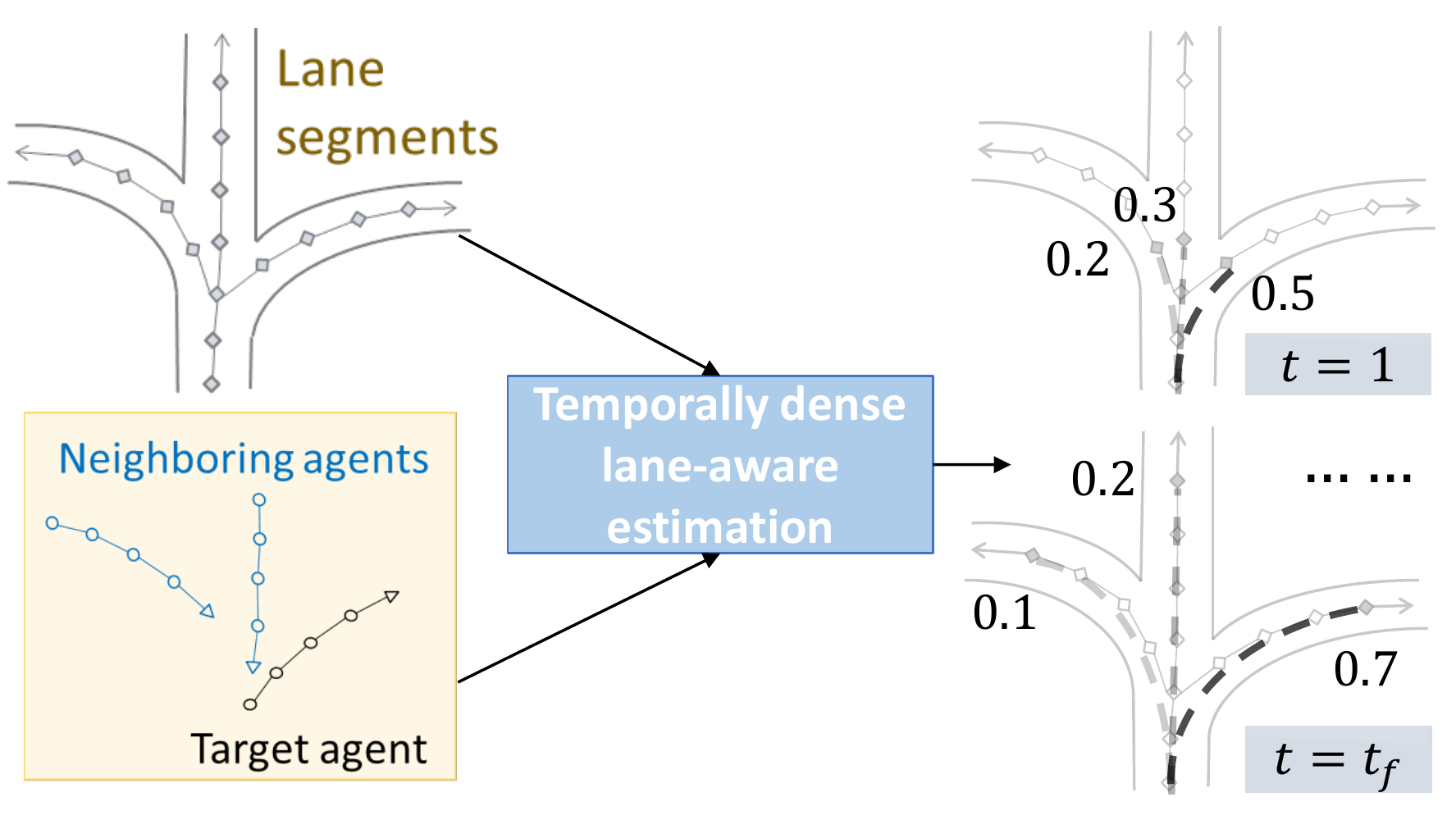}
  \caption{A temporally dense lane-aware estimation module estimates the step-wise likelihood of lane segments aligning with motion dynamics, \eg, selecting the most scene-compliant lane segments at each future time step to facilitate the decoding process.}
  \label{fig:example}
\end{figure}

Numerous data-driven approaches have been developed to tackle trajectory prediction by extracting motion dynamics from sequential trajectories and scene contexts from rasterized map data, and then fusing them in a latent space as the input to a multimodal decoder, as demonstrated in works such as Trajectron++ \cite{salzmann2020trajectron++}, CoverNet~\cite{phan2020covernet}, and AgentFormer \cite{yuan2021agentformer}. However, these approaches fail to utilize spatial and temporal information at an early stage for the subsequent decoding module. 
Moreover, rasterized map requires large receptive filters and computational cost to perceive the scene context, which may not provide accurate road structure features at complex intersections, especially for vehicle trajectory prediction.  
Consequently, the decoder may generate trajectory predictions that are non-compliant with the scene.
To mitigate this problem, VectorNet~\cite{gao2020vectornet} proposes to unify trajectory and high-definition (HD) map data into a consistent vectorized form.
This vectorization enables trajectories and lane segments based on HD maps to be easily processed and fused using the same encoder.

There are already quite a lot attempts to explore lane segments, including deep feature fusion, \eg~\cite{kim2021lapred,zeng2021lanercnn} and heuristic searching \eg~\cite{deo2022multimodal}.
We further categorize the mainstream methods into \textit{spatially} and \textit{temporally} dense methods.
Most of the current methods belong to the former one, which estimate dense probabilistic goal candidates \cite{zhao2021tnt,gu2021densetnt}, segment proposals for endpoints~\cite{wang2022ltp,gilles2021thomas,gilles2022gohome} or for the whole sequence encoding~\cite{deo2022multimodal} projected on the given scene.
We argue that these methods are suboptimal because compound prediction errors could occur if the prediction is inaccurate at initial steps.
In contrast, the temporally dense methods seek to estimate the likelihood of motion states aligning with lane positions at each time step.
Hence, the decoder has a better chance to adjust its predictions if the motion states and lane segments deviate as the time unfolds. 
However, this is not trivial because the estimation module needs to account for the variability of lane segments and uncertainty of motion states. 
Also, the alignment simply based on distance metrics~\cite{gilles2022gohome} is insufficient when the ego vehicle is at an intersection with multiple parallel lanes, or when the ego vehicle makes a lane change or a turn. 
Nevertheless, not much research has been done in exploring the temporally dense methods and not much attention has been paid to selectively feed the map information to the decoder in order to facilitate the decoding process.

To this end, we propose a temporally dense method, called LAformer.
The essence of LAformer is illustrated in Fig.~\ref{fig:example}.
It utilizes a lane-aware estimation module to select only the top-$k$ highly potential lane segments at each time step, which effectively and continuously aligns motion dynamics with scene information.
Specifically, we employ an attention-based encoder, termed Global Interaction Graph (GIG), to extract spatial-temporal features from the unified vectorized trajectories and HD map.
Different from the spatially dense methods such as \cite{wang2022ltp,gilles2021thomas, gilles2022gohome,deo2022multimodal}, we train a binary classifier using the lane information extracted from the GIG module and the target agent's motion including speed and orientation information for step-wise lane selection throughout the prediction time horizon.
Then, we introduce a Laplacian mixture density network (MDN) to generate scene-compliant multimodal trajectory predictions aligned with only the selected lane segments. 
In this way, irrelevant lane segments are filtered out to lessen the representation requirements for the decoding process. 

Additionally, to further exploit the temporal consistency over the complete time horizon, we introduce a motion refinement module.
LAformer utilizes predictions from the first stage as anchor trajectories, which distinguishes itself from anchor-points-based trajectory prediction methods using predefined anchor points \cite{chai2019multipath,varadarajan2022multipath++}.
Then the second-stage motion refinement module takes as input both the observed and predicted trajectories to further reduce prediction offsets, which is different from that of the first stage.
Although, the spirit of this refinement module is not particularly new in computer vision tasks, to our best knowledge, we are the first to effectively apply it to improve trajectory prediction. 

Our key contributions are summarized as follows:
\begin{itemize}
    \item We propose a novel temporally dense lane-aware selection method to identify the top-$k$ highly potential lane segments at each predicted time step, which is different from previous spatially dense approaches. This selection method facilitates the lane-conditioned decoder for trajectory prediction.
    \item We leverage the predicted trajectories from the first stage as anchor trajectories and introduce a second-stage motion refinement module that considers both observed and predicted trajectories. The refinement module further explores the temporal consistency across the past and future time horizons.
    \item We demonstrate the effectiveness of LAformer on two benchmark datasets, \ie,~Argoverse 1~\cite{Argoverse2019} and nuScenes~\cite{caesar2020nuscenes}. It achieves excellent performances on both benchmarks and shows superior generalized performance for the multimodal motion prediction task.
\end{itemize}

\begin{figure*}[hbpt!]
\begin{center}
 \includegraphics[clip=true, trim=0in 4.55in 0.7in 0in, width=1\linewidth]{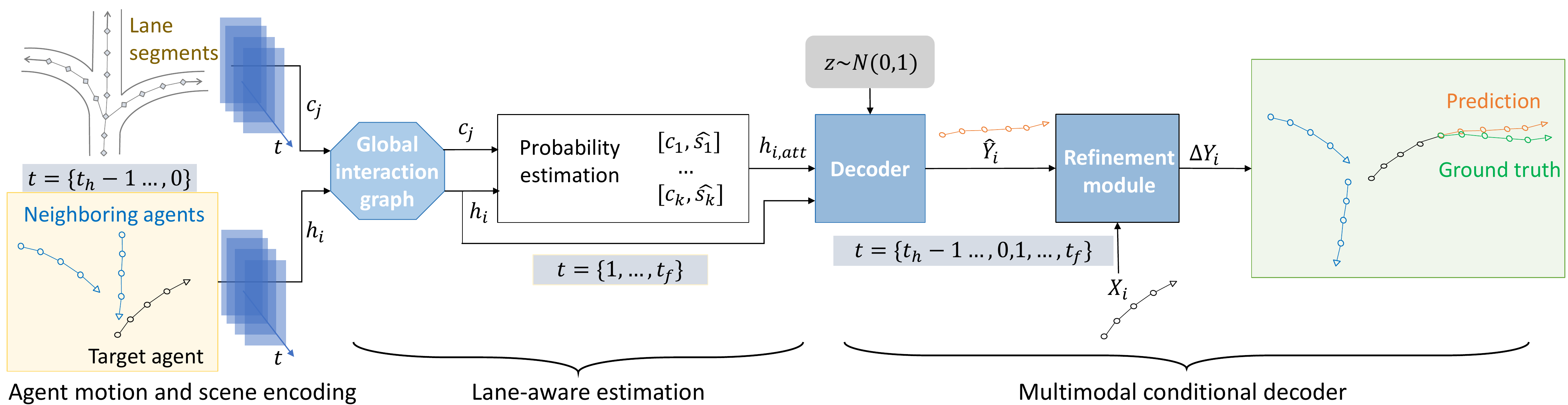}
\end{center}
  \caption{The LAformer framework takes vectorized trajectories and HD map lane segments as input. The agent motion and scene encodings are represented as $h_i$ and $c_j$, respectively, and are later fused by the attention-based global interaction graph. The decoder then takes the target agent's past trajectory $h_i$, the updated motion information aligned with the candidate lane information $h_{i, \text{att}}$ from the lane-aware estimation module, and a random latent variable $z$ as inputs, and predicts the future trajectory $\hat{Y}_{i}$. The refinement module further reduces the predicted offset $\Delta Y$ to improve prediction accuracy.}
\label{fig:framework}
\end{figure*}

\section{Related Work}
\label{sec:relatedwork}
\noindent
\textbf{Modeling interactions between agents.}
Agents are interconnected for social connections and collision avoidance~\cite{pellegrini2009you,helbing1995social}. 
Most deep learning models, \eg~\cite{alahi2016social,lee2017desire,gupta2018social,sadeghian2019sophie,salzmann2020trajectron++,yuan2021agentformer}, use agents' hidden states to aggregate the interaction information. 
The most popular aggregation strategies include pooling~\cite{alahi2016social,deo2018convolutional,gao2020vectornet}, message passing~\cite{zhang2019sr,yu2020spatio,zeng2021lanercnn} using graph convolutional networks (GCNs)~\cite{welling2017semi}, and attention mechanisms~\cite{yuan2021agentformer,liu2021multimodal,chen2022scept}. To differentiate the impacts of surrounding agents based on their relative positions and attributes, we propose the use of attention mechanisms for interaction modeling in this work.

\vspace{6pt}
\noindent
\textbf{Predicting multimodal trajectories.}
In the context of trajectory prediction for autonomous driving, predicting diverse multimodal trajectories is more favorable than single-modal trajectories to cope with agents' uncertain behaviors and scene constraints. 
Generative models, \eg,~Generative Adversarial Nets (GANs)~\cite{goodfellow2020generative}, Variational Auto-Encoder (VAE)~\cite{kingma2014auto} and conditional-VAE~\cite{kingma2014semi}, and Flows~\cite{rezende2015variational}, use sampling-based approaches to generate multiple predictions~\cite{gupta2018social,lee2017desire,rhinehart2019precog,choi2022hierarchical}. 
However, they do not provide a straightforward estimation of the likelihood of each mode. 
Although Gaussian Mixture Density Networks (MDNs) can provide a probability density function to learn the mode distribution, similar to the generative models, they often suffer from the so-called mode-collapse problem~\cite{richardson2018gans} when only a single ground truth trajectory is used for supervised learning. 
To mitigate the mode-collapse problem, this paper explores the use of a Laplacian MDN with a winner-takes-all strategy~\cite{makansi2019overcoming,zhou2022hivt,deo2022multimodal}.
Additional, in order to increase modality diversity, some approaches generate a plethora of predictions and employ ensembling techniques such as clustering or Non Maximum Suppression to reduce the predictions into a limited number of modalities~\cite{varadarajan2022multipath++,wang2022ltp}. Nevertheless, this ensembling process is time-consuming and impractical for real-time autonomous vehicles~\cite{zhou2022hivt}. Therefore, this paper refrains from adopting this technique.

\vspace{6pt}
\noindent
\textbf{Extracting scene contextual information}.
To predict scene-compliant trajectories, scene contexts must be considered.
Convolutional neural networks (CNNs) are commonly used to extract scene contexts from bird's-eye view images, such as RGB images with general contexts~\cite{lee2017desire, sadeghian2019sophie} and semantic maps with different scene categories~\cite{phan2020covernet, salzmann2020trajectron++, yuan2021agentformer}.
However, CNNs struggle to capture fine-grained scene information like lane geometry and traffic regulations.
Furthermore, the sparse information on rasterized data leads to less computational efficiency, requiring a powerful fusion module to align heterogeneous motion and scene information for the prediction module.
To address these challenges, a unified vectorization scheme~\cite{gao2020vectornet} can be used to align trajectories and lanes from an HD map.
Both trajectories and scene contexts, denoted by points, polylines, and polygons, are coded in a unified vector with coordinate information and various agent or lane attributes~\cite{liang2020learning, gu2021densetnt, zhou2022hivt, chen2022scept, ngiam2022scene, deo2022multimodal, wang2022ltp}.
This data representation is adopted in our scene-aware trajectory prediction approach.

Several approaches have been proposed to leverage lane-based scene information to guide the prediction process.
Proposal-based models~\cite{deo2018convolutional, song2020pip} classify an agent's maneuvers and then predict subsequent trajectories accordingly.
Goal-based models predict feasible goals~\cite{rhinehart2019precog,zhao2021tnt,gu2021densetnt, gilles2021thomas, gilles2022gohome} that lie in plausible lanes, and then generate complete trajectories.
Other methods use a fixed set of anchors corresponding to trajectory distribution modes to regress predicted multimodal trajectories~\cite{chai2019multipath, varadarajan2022multipath++}.
Alternatively, \cite{zhong2022aware} proposes a method that treats the collection of historical trajectories at an agent's current location as prior information to narrow down the search space for potential future trajectories.
We categorize these methods as spatially dense lane-based methods, as they focus on generating a probabilistic distribution of candidate goals or full trajectories over the space.
However, these methods do not fully explore temporal information to account for motion uncertainty and scene variability as time progresses. Additionally, the prediction module must implicitly filter out irrelevant scene information, which can be challenging in complex scene constraints, such as those present at intersections.

In contrast to these methods, we propose a temporally dense lane-aware module to learn the alignment between an agent's motion dynamics and potential lane segments.
Instead of simply combining motion encoding and lane encoding and allowing the decoder to implicitly learn their relationship~\cite{park2023leveraging}, we explicitly estimate the likelihood of the lane that an agent will take at each time step.
Then, we select only the top highly potential lane information to balance the variability of lane segments and uncertainty of motion dynamics.

\section{Method}
\label{sec:methodology}

\subsection{Problem formulation}
\label{sub:problemformulation}
Following the mainstream works, \eg.~\cite{gao2020vectornet,gu2021densetnt,wang2022ltp,zhou2022hivt,deo2022multimodal}, we assume that detecting and tracking road agents, as well as perceiving the environment, provides high-quality trajectory and HD map data in a 2D coordinate system. 
Namely, for agent $i$, the $x$- and $y$-positions within a given time horizon $\{-{t_{h}}+1, \cdots, 0, 1,\cdots, t_f\}$ are obtained, along with the HD map $\mathbf{C}$ of the scene. 
The downstream task is to predict the subsequent trajectories $\mathbf{Y}_{1:t_f}^{i}$ by leveraging the HD map and observed trajectories of all agents in the given scenario, including the target agent's trajectory $\mathbf{X}_{-t_h+1:0}^{i}$.

Both agents' past trajectories and lane centerlines are represented as vectors. 
To be more specific, for agent $i$, its history trajectory $X_i$ is represented as an ordered sequence of sparse trajectory vectors $\mathbf{A}^{i}_{-t_{h}+1:0}=\{v^{i}_{-t_{h}+2},v^{i}_{-t_{h}+3},...,v^{i}_{0}\}$ over the past $t_{h}$ time steps. 
Each trajectory vector $v^{i}_{t}$ is defined as $v^{i}_{t}=[d^{i}_{t, s}, d^{i}_{t, e}, a^{i}]$, where  $d^{i}_{t, s}$ and $d^{i}_{t, e}$ denote the start and end points, respectively, and $a_{i}$ corresponds to agent $i$'s attribute features, such as timestamp and object type (\ie, autonomous vehicles, target agent, and others).
In addition, lane centerlines are further sliced into predefined segments to capture fine-grained lane information in order to model an agent's intention precisely. Similar to the trajectory vector, a lane centerline segment is represented as $\mathbf{C}^{i}_{1:N}=\{v^{i}_{1},v^{i}_{2},...,v^{i}_{N}\}$, where $N$ denotes the total vector length. Each lane vector $v^{i}_{n}=[d^{i}_{n, s}, d^{i}_{n, e}, a_{i}, d^{i}_{n, \text{pre}}]$ adds $d^{i}_{n, \text{pre}}$ to indicate the predecessor of the start point. The lane vectors are connected end-to-end to obtain the HD map's structural features.

Moreover, to ensure input feature invariance with respect to an agent's location, the coordinates of all vectors are normalized to be centered around the target agent's last observed position. 

Fig.~\ref{fig:framework} presents the overall framework of LAformer, which takes vectorized trajectories and HD map lane segments as input and outputs multimodal trajectories for the target agent.
Each module of LAformer is explained in detail below. 

\subsection{Agent motion and scene encoding}
\label{subsec:scene_encoder}
We design an attention-based Global Interaction Graph (GIG) to encode agent motion and scene information.
Concretely, we process trajectory vectors $\mathbf{A}^{i}$ and lane vectors $\mathbf{C}^{j}_{1:N}$ using a Multi-Layer Perceptron (MLP) and a Gated Recurrent Unit (GRU) layer in a sequential manner.
The output encodings of these layers are represented as $h_i$ for $\forall i \in \{1,\dots, N_\text{traj}\}$ and $c_j$ for $\forall j \in \{1,\dots, N_\text{lane}\}$ in a given scenario.   
To fuse these encodings, we design a symmetric cross attention mechanism that operates on $h_i$ and $c_j$ as follows: 
\begin{align} 
\label{eq:hccross}
h_i &= h_i + \operatorname{CrossAtt}\{h_i, c_j\}~\text{for}~j \in \{1,\dots, N_\text{lane}\},\\
c_j &= c_j + \operatorname{CrossAtt}\{c_j, h_i\}~\text{for}~i \in \{1,\dots, N_\text{traj}\}.
\end{align}
Afterwards, the GIG further explores the self-attention and skip-connection to learn the interactions among agents.
\begin{align}
    h_i &= \operatorname{ConCat}[h_i, c_j]~\text{for}~j \in \{1,\dots, N_\text{lane}\},\\  
    h_i &= h_i + \operatorname{SelfAtt}\{h_i\}~\text{for}~i \in \{1,\dots, N_\text{traj}\}. 
\end{align}

\subsection{Temporally dense lane-aware estimation}
\label{subsec:dense_lane_aware}
We propose a temporally dense lane-aware probability estimation module that uses attention to guide a target agent towards the most influential lane segments for its future trajectories. 
Specifically, we align the target agent's motion and lane information at each future time step $t \in \{1, \dots, t_f\}$.
To achieve this, we predict lane probabilities using a lane scoring header and an attention mechanism.
The key ($K$) and value ($V$) vectors are linear projections of the agent motion encoding $h_{i}$, while the query ($Q$) vector is a linear projection of the lane encoding $c_{j}$. 
These vectors are then fed into a scaled dot-product attention block $A_{i,j} = \text{softmax}(\frac{QK^{T}}{\sqrt{d_{k}}})V$, resulting in the predicted score of the $j$-th lane segment at $t$ given by
\begin{equation}
\label{eq:softmax_dense}
\hat{s}_{j,t} = \frac{\exp({\phi}\{h_i, c_j, A_{i,j}\})}{\sum^{N_\text{lane}}_{n=1} \exp({\phi}\{h_i, c_n, A_{i,n}\})},
\end{equation}
where ${\phi}$ denotes a two-layer MLP.

To balance the variability of lane segments and uncertainty of motion dynamics, we select the top-$k$ lane segments $\{c_{1},c_{2},\dots,c_{k}\}$ with the $k$ highest scores $\{\hat{s}_{1},\hat{s}_{2},\dots,\hat{s}_{k}\}$ as the candidate lane segments. We then concatenate the candidate lane segments and associated scores over the future time steps to obtain $C=\operatorname{ConCat}\{c_{1:k}, \hat{s}_{1:k}\}_{t=1}^{t_f}$.
Next, we perform cross attention to project the target agent's past trajectory encoding $h_i$ as the query vector, and the candidate lane encodings $C$ as the key and value vectors. The output is updated motion information aligned with the lane information, denoted as $h_{i, \text{att}}$.
This cross attention further explores scene information in spatial and temporal dimensions.

The lane scoring module uses a binary cross-entropy loss $\mathcal{L}_\text{lane}$ to optimize the probability estimation. The ground truth value $s_t$ is set to 1 for the lane segment that is closest to the trajectory' truth position, and 0 for all other lanes. It is worth mentioning that the ground truth lane segment $s_t$ does not need additional labeling and can be identified easily using a distance metric, such as the Euclidean distance.

\begin{equation}
\label{eq:loss_lane}
\mathcal{L}_{lane} = \sum^{t_\text{f}}_{t=1}\mathcal{L}_\text{CE}({s}_{t},\hat{s}_{t}).
\end{equation}

\subsection{Multimodal conditional decoder}
\label{subsec:decoder}
This section introduces a Laplacian mixture density network (MDN) decoder that is conditioned on the encodings of the target agent's past trajectory $h_i$ and the updated motion information aligned with the candidate lane information $h_{i, \text{att}}$. 
To further preserve the diversity of multimodalities, we sample a latent vector $z$ from a multivariate normal distribution, which serves as an additional condition added to the encodings for the predictions. 
The decoder predicts a set of trajectories $\sum^{M}_{m=1}\hat{\pi}_{m}\,\text{Laplace}(\mu,\,b)$, where $\hat{\pi}_{m}$ denotes the probability of each mode indexed by $m$ among the $M$ predicted modes and $\sum^{M}_{m=1}\hat{\pi}_{m}=1$. 
Here, $\mu$ and $b$ represent the location and scale parameters of each Laplace component.
We use an MLP to predict $\hat{\pi}_{m}$, 
a GRU to recover the time dimension $t_{f}$ of the predictions, and two side-by-side MLPs to predict $\mu$ and $b$.

We train the Laplacian MDN decoder by minimizing a \textit{regression loss} and a \textit{classification loss}. The regression loss is computed using the Winner-Takes-All strategy~\cite{makansi2019overcoming,zhou2022hivt,deo2022multimodal} and is defined as:
\begin{equation}
 \mathcal{L}_{\text{reg}}=\frac{1}{t_{f}}\sum^{t_{f}}_{t=1}-\log P({Y_{t}}|\mu^{m^*}_{t}, \mathbf{b}^{m^*}_{t}),
\end{equation}
where $Y$ represents the ground truth position and $m^*$ represents the mode with the minimum $L_2$ error among the $M$ predictions. 
The cross-entropy loss is used to optimize the mode classification and is defined as:
\begin{align}
\mathcal{L}_{\text{cls}}=\sum^{M}_{m=1}-\pi_{m}\log(\hat{\pi}_{m}).
\end{align}
We adopt the soft displacement error, following~\cite{zhou2022hivt}, as our target probability $\pi_{m}$. 
The total loss for the motion prediction in the first stage is given by:
\begin{equation}
\label{eq:s1loss}
\mathcal{L}_{\text{S1}}=\lambda_{1}\mathcal{L}_{\text{lane}}+\mathcal{L}_{\text{reg}}+\mathcal{L}_{\text{cls}},
\end{equation}
where $\lambda_{1}$ controls the relative importance of $\mathcal{L}_{\text{lane}}$.

\subsection{Motion refinement}
\label{subsec:refinement}
A second-stage motion refinement is introduced to further explore the temporal consistency for predicting more accurate future trajectories.
The goal is to reduce the offset between ground truth trajectory $Y_{1:t_{f}}$ and predicted trajectory $\hat{Y}_{1:t_{f}}$.
In this stage, we leverage the complete trajectory $\{\{X\}_{-t_h+1}^0, \{\hat{Y}\}_{1}^{t_f}\}$ as the input to extract the motion encoding $\dot{h}_i$ using a similar temporal encoder as in the first stage. 
Then, a regression header constructed by a two-layer MLP takes as input all the motion encodings $[h_i, h_{i, \text{att}}, \dot{h}_i]$ in both stages and predicts the offset $\Delta{Y} = Y - \hat{Y}_{m}$ between the ground truth and predicted trajectories.
We use $L_2$ loss to optimize the offset. 
\begin{equation}
\label{eq:offsetloss}
\mathcal{L}_\text{off}= \frac{1}{t_{f}}\sum^{t_{f}}_{t=1}||\Delta\hat{Y}_{t} - \Delta{Y}_{t}||_2.
\end{equation}

Furthermore, we use a cosine function, denoted by Eq.~\eqref{eq:poseerror}, to explicitly aid the model in learning the turning angle from the last observed position. 
It measures the difference between the ground truth angle $\theta_{t}= \text{arctan2}(Y_{t}-{X}_{0})$ and the predicted angle $\hat{\theta}_{t}= \text{arctan2}(\hat{Y}_{t}-{X}_{0})$.
\begin{equation}
\label{eq:poseerror}
    \mathcal{L}_{\text{angle}}=\frac{1}{t_{f}}\sum^{t_{f}}_{t=1}-cos(\hat{\theta}_{t}-\theta_{t}).
\end{equation}
Here, we employ a Winner-Takes-All strategy to optimize the offset and angle losses, similar to the first stage. 
The total loss in the second stage can be expressed as: 
\begin{equation}
\label{eq:s2loss}
\mathcal{L}_{\text{S2}}=\mathcal{L}_{\text{S1}}+\lambda_{2}\mathcal{L}_{\text{off}}+\lambda_{3}\mathcal{L}_{\text{angle}},
\end{equation}
where $\lambda_{2}$ and $\lambda_{3}$ control the relative importance of the corresponding loss terms.

\section{Experiments}
\label{sec:experiments}
\subsection{Experimental setup}
\label{subsec:setup}
\noindent
\textbf{Datasets.} 
The proposed approach is developed and evaluated on two challenging and widely used benchmarks for autonomous driving: \textit{nuScenes}~\cite{caesar2020nuscenes} and \textit{Argoverse 1}~\cite{Argoverse2019}. 
These benchmarks provide trajectories of various types of road agents with an HD map of the given scene. 
In nuScenes, the target agent's subsequent six-second trajectory is predicted based on its and neighboring agents' trajectories up to two seconds, with trajectory sampling at \SI{2}{Hz}. 
In Argoverse 1, the target agent's subsequent three-second trajectory is predicted based on its and neighboring agents' trajectories in the initial two seconds, with trajectory sampling at \SI{10}{Hz}. 
To ensure a fair comparison, the official data partitioning and online test server of both benchmarks are used for the training and test setting, respectively.

\vspace{3pt}
\noindent\textbf{Evaluation metrics:}
We adopt the standard evaluation metrics to measure prediction performance, including $\text{FDE}_\mathsf{K}$ and $\text{ADE}_\mathsf{K}$ for measuring $L_2$ errors at the final step and averaged at each step, respectively, for predicting $\mathsf{K}$ modes.
Here, the minimum error of the $\mathsf{K}$ modes is reported. 
Both ADE and FDE are measured in meters.
In addition, the miss rate $\text{MR}_\mathsf{K}$ measures the percentage of scenarios for which the final-step error is larger than \SI{2.0}{m}.
$\mathsf{K}$ is set to 5 and 10 in nuScenes and 6 in Argoverse 1 for the multimodal trajectory prediction.
For all the evaluation metrics, the lower the better.

\vspace{3pt}
\noindent
\textbf{Implementation details.} 
The hidden dimension of all the feature vectors in LAformer is set to 128. 
Only the lane segments that are within \SI{50}{m} (Manhattan distance) of the target agent are sampled as the scene contexts. 
$\lambda_1$ in Eq.~\eqref{eq:s1loss} is set to 10.
$\lambda_2, \lambda_3$ in Eq.~\eqref{eq:s2loss} are set to 5 and 2, respectively.
We use a two-stage training scheme. 
In the first stage, all the modules except for the motion refinement module are trained using the Adam optimizer~\cite{kingma2015adam}.
In the second stage, all the modules are trained together.
LAFormer was trained on 8xRTX3090 cards with each stage for about 8 hours~\footnote{More details about the implementation and training scheme can be found in the supplementary material.}. 

\subsection{Quantitative results and comparison}
\label{subsec:results}
\begin{table}[ht!]
\centering
\begin{tabular}{l|cc|cc}
\toprule
\multirow{2}{*}{Paper}  & \multicolumn{2}{c}{Val set} & \multicolumn{2}{c}{Test set} \\ 
      & ADE & FDE & ADE & FDE \\ \midrule
{THOMAS \cite{gilles2021thomas}}{\color{red}$\ddagger$} &-&-& 0.94 & 1.44 \\
TNT \cite{zhao2021tnt} & 0.73 & 1.29 & 0.91 & 1.45\\
LaneRCNN \cite{zeng2021lanercnn} & 0.77 & 1.19 & 0.90 & 1.45\\
{GOHOME \cite{gilles2022gohome}}{\color{red}$\ddagger$} &-&-& 0.89 & 1.29 \\
DenseTNT \cite{gu2021densetnt} & 0.73 & 1.05 & 0.88 & 1.28\\
LaneGCN \cite{liang2020learning} & 0.71 & 1.08 & 0.87 & 1.36\\
{LaneGCN \cite{liang2020learning,zhong2022aware}}$\diamondsuit${\color{red}$\ddagger$} & - &-&0.84 & 1.30\\
mmTrans \cite{liu2021multimodal} &-&-& 0.84  & 1.34 \\
MultiModalTrans \cite{huang2022multi} & & & 0.84 & 1.29 \\
LTP \cite{wang2022ltp} &-&-& 0.83 & 1.29 \\
TPCN \cite{ye2021tpcn} & 0.73 & 1.15 & 0.82 & 1.24 \\
{FRM \cite{park2023leveraging}}{\color{red}$\ddagger$} & 0.68 & 0.99 & 0.82 & 1.27 \\
SceneTrans \cite{ngiam2022scene} &-&-& 0.80 & 1.23 \\
Multipath++ \cite{varadarajan2022multipath++} &-&-& \underline{0.79} & 1.21 \\
HiVT \cite{zhou2022hivt} & \underline{0.66} & \underline{0.96} & \textbf{0.77} & \underline{1.17} \\
{LAformer (Ours)}{\color{red}$\ddagger$} & \textbf{0.64} & \textbf{0.92} &\textbf{ 0.77} & \textbf{1.16} \\
\bottomrule
\end{tabular}
\caption{The results on \textit{Argoverse 1} \cite{Argoverse2019} validation set and the online test set with $\mathsf{K}=6$. $\diamondsuit$: the model is enhanced by local behavior data LBA~\cite{zhong2022aware}. {\color{red}$\ddagger$}: the models tested on both benchmarks. The best/second-best values are highlighted in boldface/underlined.}
\label{tab:resultsonargoverse}
\end{table}

\begin{table}[ht!]
\centering
\begin{tabular}{l|cc|cc}
\toprule
\multirow{2}{*}{Paper}           & \multicolumn{2}{c|}{$\mathsf{K} = 5$} & \multicolumn{2}{c}{$\mathsf{K} = 10$} \\
                                 & ADE      &  MR      &  ADE      &  MR
                                 \\ \midrule
Multipath \cite{chai2019multipath} &2.32 &- &1.96 &- \\                   
CoverNet \cite{phan2020covernet} &1.96 &0.67 &1.48 &- \\
Trajectron++ \cite{salzmann2020trajectron++} & 1.88 &0.70& 1.51& 0.57\\
AgentFormer \cite{yuan2021agentformer} & 1.86 &-& 1.45& -\\
ALAN \cite{narayanan2021divide} & 1.87 &0.60 &1.22 &0.49 \\
SG-Net \cite{wang2022stepwise} & 1.86& 0.67& 1.40& 0.52 \\
WIMP \cite{khandelwal2020if} & 1.84 &{0.55} &1.11 &0.43 \\
MHA-JAM \cite{messaoud2020multi} & 1.81 &0.59 &1.24 &46 \\
CXX \cite{luo2020probabilistic} &1.63 &0.69 &1.29 &0.60 \\
LaPred \cite{kim2021lapred} & 1.53&-& 1.12 &-\\
P2T \cite{deo2020trajectory} &1.45 &0.64 &1.16& 0.46\\
{LaneGCN \cite{liang2020learning,zhong2022aware}}$\diamondsuit${\color{red}$\ddagger$} & - & \underline{0.49} & 0.95 & 0.36 \\ 
{GOHOME \cite{gilles2022gohome}}{\color{red}$\ddagger$} & 1.42 &0.57 &1.15 &0.47 \\
Autobot \cite{girgis2021latent} &1.37 &0.62 &1.03 &0.44 \\
{THOMAS \cite{gilles2021thomas}}{\color{red}$\ddagger$} &1.33 &{0.55} &1.04 &- \\
PGP~\cite{deo2022multimodal} & 1.30 &0.61 &1.00 &\underline{0.37}\\
{FRM \cite{park2023leveraging}}{\color{red}$\ddagger$}  & \textbf{1.18} & \textbf{0.48} & \textbf{0.88} & \textbf{0.33} \\ 
{LAformer (Ours)}{\color{red}$\ddagger$}   & \underline{1.19} & \textbf{0.48} & \underline{0.93} & \textbf{0.33}\\ 
\bottomrule                                              
\end{tabular}
\caption{The results on \textit{nuScenes} \cite{caesar2020nuscenes} online test set. $\diamondsuit$: the model is enhanced by local behavior data LBA~\cite{zhong2022aware}. {\color{red}$\ddagger$}: the models tested on both benchmarks. The best/second best values are highlighted in boldface/underlined.}
\label{tab:resultsonnuscenes}
\end{table}

Tables~\ref{tab:resultsonargoverse} and \ref{tab:resultsonnuscenes} present the results obtained on the \textit{Argoverse 1} validation and online test sets, and the \textit{nuScenes} online test set, respectively. The leaderboard results (online tests) are updated up to 2023-02-20 according to the officially published papers.

In the Argoverse 1 benchmark, LAformer achieves the state-of-the-art performance on the validation set by a clear margin in ADE and FDE. It also achieves excellent results on the test set, on par with the runner-up method HiVT.

In the nuScenes benchmark, LAformer achieves competitive performance, only slightly inferior to the newly released FRM~\cite{park2023leveraging} in terms of ADE. FRM introduces relationship reasoning to help understand future interactions between the ego and other agents, while LAformer relies on attention mechanisms to learn interactions between agents and focuses more on scene constraints. This difference in approach may contribute to the performance difference.
However, LAformer outperforms other lane-based models, \eg,~LaneGCN \cite{liang2020learning,zhong2022aware} and PGP~\cite{deo2022multimodal}, with a clear margin, indicating that our lane-aware estimation is more effective than the other distance-based or heuristic lane searching. 

Moreover, when compared to the models (marked by {\color{red}$\ddagger$}) tested on both benchmarks, namely THOMAS~\cite{gilles2021thomas}, GOHOME~\cite{gilles2022gohome}, LaneGCN~\cite{liang2020learning}$\diamondsuit$ enhanced by local behavior data LBA~\cite{zhong2022aware}, as well as FRM, LAformer shows evidently more generalized performance across the benchmarks. This suggests that the proposed temporally dense lane-aware estimation module effectively aligns scene constraints with motion dynamics, even though the trajectories provided in Argoverse 1 and nuScenes include locations in different cities and driving directions. Further evidence supporting the efficacy of this module can be found in the following ablation study presented in Table~\ref{tab:denselane}.

\subsection{Ablation study}
\label{subsec:ablation}
Considering data scale and availability of ground truth, we carry out the ablation study on the Argoverse 1 validation set with 39,472 sequences.
The Baseline model predicts future trajectories only conditioned on the observed trajectories of the target and its neighboring agents, with the second refinement module (S2) and lane-aware estimation module removed.
LAformer (Spa.) only estimates the likelihood of goal position aligning with the lane information, similar to the spatially dense models.
In contrast, LAformer (Tem.) estimates the likelihood of the position at each time step aligning with the temporally dense lane information.
LAformer (Full) is the complete proposed model.
\begin{table}[ht!]
\centering
\begin{tabular}{l|ccc|cc}
\toprule
Name & S2       & G & D & $\text{ADE}_6$  & $\text{FDE}_6$   \\ \midrule
Baseline& -        & -  & -  & 0.72 &  1.12 \\
Baseline+S2 & $\surd$  & -  & -  & 0.71  & 1.08   \\ 
LAformer (Spa.) & -     & $\surd$ & -  & 0.69 &  1.03 \\ 
LAformer (Tem.) & -        & -     & $\surd$ & 0.66 &  0.95\\
LAformer (Full) & $\surd$   & -     & $\surd$ & 0.64 & 0.92 \\\bottomrule
\end{tabular}
\caption{Ablation study on the lane estimation and refinement modules. \textbf{S2}: the second stage refinement module, \textbf{G}: only the goal (last step) of the lane segments are scored, and \textbf{D}: the temporally dense lane-aware module.} 
\label{tab:denselane}
\end{table}

From Table~\ref{tab:denselane}, the performance of the Baseline is much inferior to the other models.
The comparison of Baseline vs. Baseline+S2 and LAformer (Tem.) vs. LAformer (Full) demonstrates the performance gain of S2, \eg,~ca. 3\% in FDE.
The comparison of LAformer (Spa.) vs. LAformer (Tem.) shows that our temporally dense method is more effective than the spatially dense method, reducing ADE by about 4\% and FDE by about 8\%.

Furthermore, we carry out an ablation study to analyze the effectiveness of the angle ${\mathcal{L}_\text{angle}}$ and offset $\mathcal{L}_{\text{off}}$ losses in the second stage. 
As shown in Table~\ref{tab:loss2}, by including both losses helps to improve the prediction accuracy by about 2\% measured in ADE and FDE.
\begin{table}[ht]
\centering
\begin{tabular}{cc|cc}
\toprule
${\mathcal{L}_\text{angle}}$ & $\mathcal{L}_{\text{off}}$ & $\text{ADE}_6$  & $\text{FDE}_6$   \\ \midrule
$\surd$      & -       & 0.65 &  0.94\\
 -      & $\surd$       & 0.65 &  0.92 \\
$\surd$      & $\surd$       & 0.64  & 0.92\\ \bottomrule
\end{tabular}
\caption{Ablation study on the loss functions to penalize the angle ${\mathcal{L}_\text{angle}}$ and offset $\mathcal{L}_{\text{off}}$ errors in the second stage.} 
\label{tab:loss2}
\end{table}

We also ablate the latent variable $z$ added to the input of the multimodal conditional decoder.
It is sampled from a multivariate normal distribution with its dimension setting to 2.
However, as shown in Table~\ref{tab:z}, we find that inserting $z$ only leads to a marginal performance improvement (less than \SI{0.5}{cm}).
\begin{table}[ht]
\centering
\begin{tabular}{l|cc}
\toprule
 $z$ & $\text{ADE}_{6}$ & $\text{FDE}_{6}$ \\ \midrule
 - & 0.692 &  1.035 \\
 $\surd$ & 0.690 & 1.032 \\ \bottomrule
\end{tabular}
\caption{Ablation study on the latent variable $z$.}
\label{tab:z}
\end{table}

\subsection{Sensitivity analysis of the hyper-parameters}
\label{subsec:sensitivity}
The number of top-$k$ lane segments and the weights of losses are crucial hyper-parameters for LAformer. To examine their impact, we conduct an empirical study by varying their values around the experimental settings indicated by an underline in Tables~\ref{tab:topk} and \ref{tab:lambdas}.

As shown on the left side of Table~\ref{tab:topk}, increasing the number of lane segments from 1 to 4 initially results in a performance gain up to $k=3$, but after that, it starts to decline. In the second stage, $k=2$ provides better results than 3. Using a larger $k$ increases the chances of including irrelevant lane segments, while a relatively small $k$ enables the decoder to focus on the most relevant lane segments.
\begin{table}[ht]
\begin{tabular}{cc}
    \begin{minipage}{.42\linewidth}
        \centering
        \begin{tabular}{c|cc}
        \toprule
        $k$ & $\text{ADE}_6$  & $\text{FDE}_6$    \\ \midrule
        1 & 0.68 & 1.00\\
        \underline{2} & 0.66 & 0.95 \\
        3 & 0.65 & 0.95 \\
        4 & 0.66 & 0.96 \\ \bottomrule
        \end{tabular}
        \label{tab:topk1}
    \end{minipage} &

    \begin{minipage}{.42\linewidth}
       \centering
        \begin{tabular}{c|cc}
        \toprule
        $k$  & $\text{ADE}_6$  & $\text{FDE}_6$    \\ \midrule
        \underline{2}        & 0.64 & 0.92\\ 
        3        & 0.64 & 0.93\\ \bottomrule       
        \end{tabular}
        \label{tab:topk2}
    \end{minipage} 
\end{tabular}
\caption{The number of top lane segments that impacts the prediction performance. Left: first stage, Right: second stage.} 
\label{tab:topk}
\end{table}

We also vary the loss weights $\lambda_{1}$ in Eq.~\eqref{eq:s1loss} and $\lambda_{2}, \lambda_{3}$ in Eq.~\eqref{eq:s2loss}. As shown in Table~\ref{tab:lambdas}, we only observe marginal performance differences, \eg, $\text{ADE}_{6}$ fluctuates within \SI{1}{cm}.
\begin{table}[ht]
    \begin{tabular}{cc}
    \begin{minipage}{.44\linewidth}
        \centering
        \begin{tabular}{c|cc}
        \toprule
         $\lambda_{1}$ & $\text{ADE}_6$ & $\text{FDE}_6$ \\ \midrule
         8 & 0.70 &  1.05\\
         9 & 0.70  & 1.02 \\
         \underline{10} & 0.69 &  1.01 \\
         11 & 0.70 &  1.05 \\
         12 & 0.70 &  1.05 \\ \bottomrule
        \end{tabular}
    \end{minipage} &

    \begin{minipage}{.44\linewidth}
       \centering
        \begin{tabular}{c|cc}
        \toprule
         $\lambda_{2}$ & $\text{ADE}_6$ & $\text{FDE}_6$ \\ \midrule
         1 &  0.68 &   1.01\\
         \underline{5} & 0.68  & 1.00   \\
         10&  0.69 &  1.02 \\
         \midrule
         $\lambda_{3}$ & $\text{ADE}_6$ & $\text{FDE}_6$ \\ \midrule
         1 & 0.68 &  1.00 \\
         \underline{2} & 0.68  & 1.00 \\
         10 & 0.68  & 1.01  \\ \bottomrule
        \end{tabular}
    \end{minipage} 
\end{tabular}
\caption{The number of top lane segments that impacts the prediction performance. Left: first stage, Right: second stage.} 
\label{tab:lambdas}
\end{table}

\subsection{Computational performance}
As reported in Table~\ref{tab:computation}, LAformer has 2,645K parameters, similar to HiVT-128 but larger than LTP and DenseTNT. Its inference time for a scenario with an average of 12 agents is ca. \SI{115}{ms}, which is not a main strength compared to HiVT. But this inference speed is comparable to LTP and faster than DenseTNT and PGP, making LAformer close to real-time use cases at \SI{10}{Hz}. 
\begin{table}[ht!]
\begin{tabular}{l|c|cc}
\toprule
\multirow{2}{*}{Model}    & \multirow{2}{*}{\#Params} & \multicolumn{2}{c}{Inference speed} \\
         &          & Batch size        & Time (ms)       \\ \midrule
LTP \cite{wang2022ltp}      &   1,100k    & 8               & 92              \\
DenseTNT \cite{gu2021densetnt} &1,103K          & 32                & 531             \\
HiVT-128 \cite{zhou2022hivt} & 2,529K   & 32                & 38              \\
PGP \cite{deo2022multimodal}      &     -     & 12                & 215             \\
LAformer (Ours) & 2,654K   & 12                & 115             \\
\bottomrule
\end{tabular}
\caption{Computational performance.}
\label{tab:computation}
\end{table}

\begin{figure*}[ht]
\centering
\includegraphics[clip=true, trim=0in 5.15in 3.75in 0.0in, width = 0.95\linewidth]{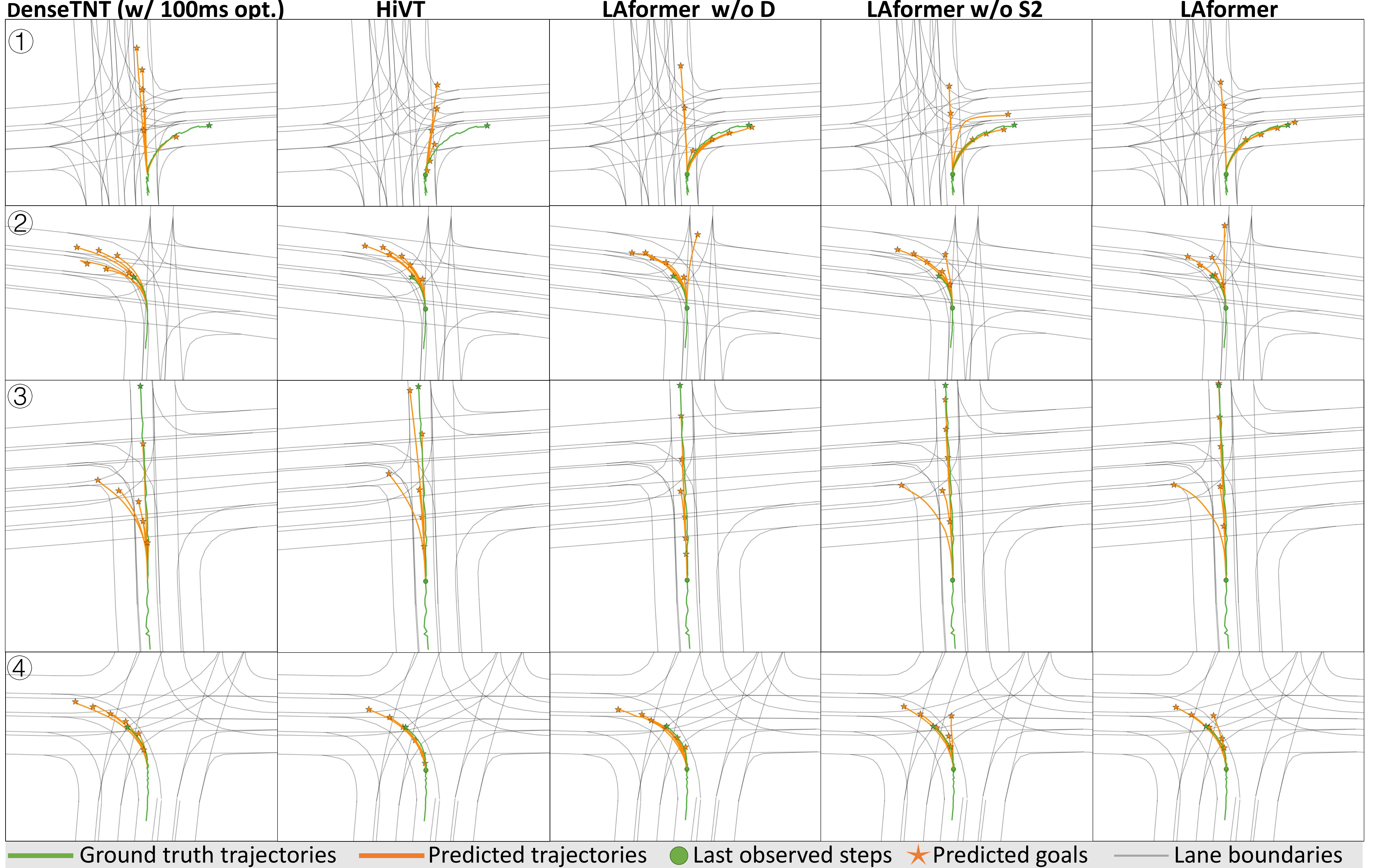}
\caption{Qualitative comparison of models in complex scenarios, with each row representing a unique intersection scenario and each column representing results predicted by the same model. Models include: LAformer w/o D (proposed model without temporally dense lane-aware module), LAformer w/o S2 (proposed model with temporally dense lane-aware module but without second-stage motion refinement), and LAformer (complete model) in the rightmost column.}
\label{fig:ablation_fig}
\end{figure*}
\begin{figure}[ht]
\centering
\includegraphics[clip=true, trim=0in 0in 3in 0.0in, width = 1\linewidth]{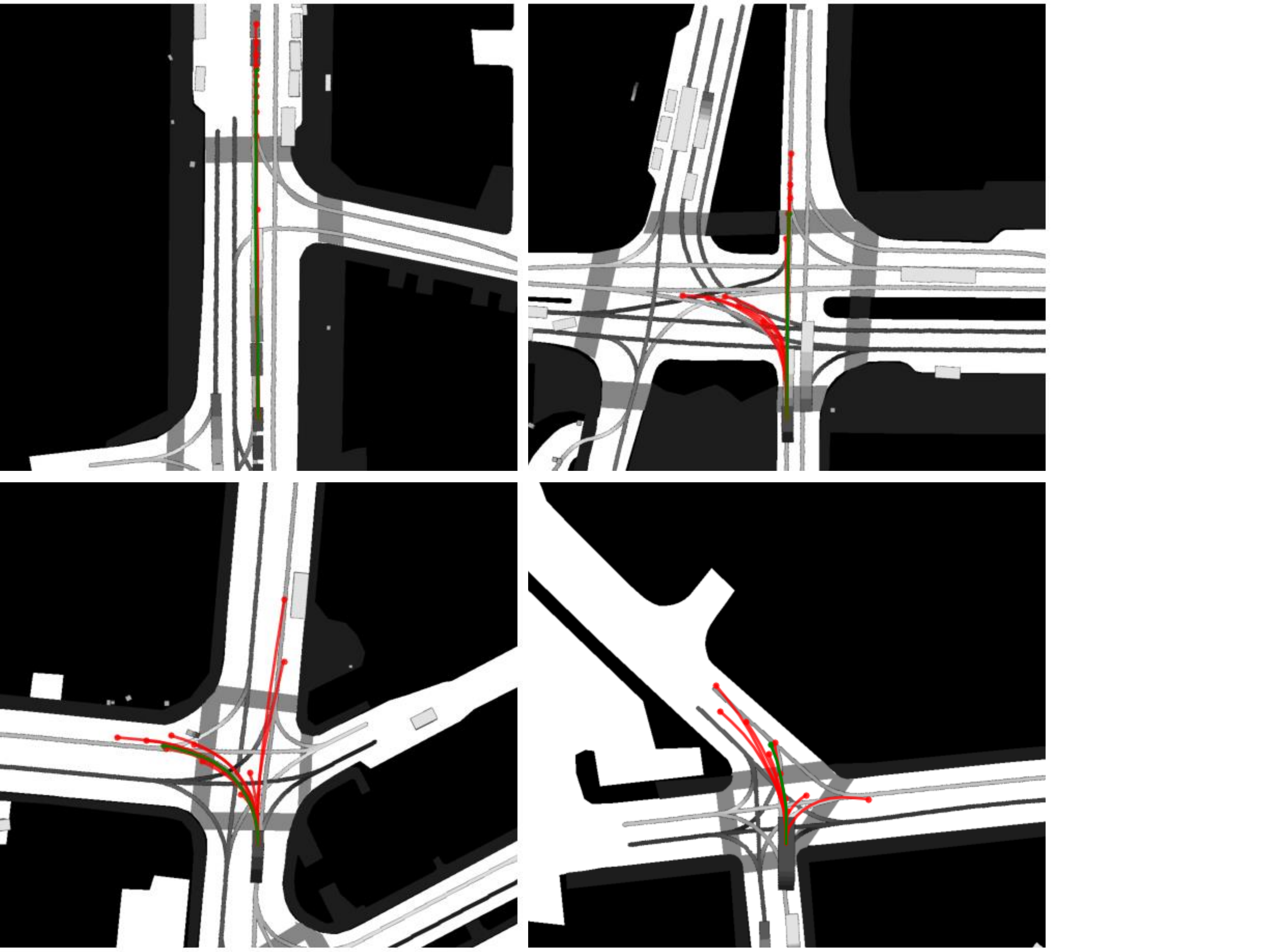}
\caption{The qualitative results of LAformer on nuScenes. Predicted trajectories are presented in red color and the corresponding ground truth trajectories are presented in green color.}
\label{fig:nuscene_vis}
\end{figure}
\subsection{Qualitative results}
Figure~\ref{fig:ablation_fig} presents qualitative results of LAFormer compared to runner-up models on the \textit{Argoverse 1} validation set~\footnote{More qualitative results are presented in the supplementary material.}. To ensure a fair comparison, we use the publicly trained HiVT~\cite{zhou2022hivt} model to replicate the results reported in their paper. As DenseTNT~\cite{gu2021densetnt} does not provide a trained model, we retrain the offline model with optimization (DenseTNT w/100ms opt.) from scratch to achieve similar performance to that reported in their paper.
All the models produce reasonable multimodal predictions for the target agent in various traffic scenarios at intersections, such as turning right \textcircled{1} and left \textcircled{2}, \textcircled{4}, or driving straight with acceleration \textcircled{3}. However, LAformer generates more accurate predictions in the right-turn scenario \textcircled{1} and the acceleration scenario \textcircled{3}, while other models tend to predict decelerating or turning modes.
Furthermore, when the temporally dense lane-aware module is deactivated (w/o D vs. w/o S2), LAformer generates less diverse predictions in the lateral directions. However, the complete model with the second-stage refinement module shown in the rightmost column maintains good prediction diversity and accuracy.

Figure~\ref{fig:nuscene_vis} presents more qualitative results of LAformer on nuScenes. It not only generates accurate prediction in straightforward driving but also at complicated intersections, for example, continuing driving forward, turning left or right. The multimodal predictions aligned with lane segments implies the agent's motion uncertain at intersections.

\section{Conclusion}
\label{sec:conclusion}
The paper presents LAformer, an end-to-end attention-based trajectory prediction model that takes observed trajectories and an HD map as input and outputs a set of multimodal predicted trajectories. A Transformer-based temporally dense lane-aware module and a second-stage motion refinement module are used to improve prediction accuracy. LAformer outperforms other models on both Argoverse 1 and nuScenes motion forecasting benchmarks, demonstrating a superior generalized performance.
Moreover, extensive ablation and sensitivity studies verify the efficacy of the lane-aware and motion refinement modules.

\newpage
{\small
\bibliographystyle{ieee_fullname}
\bibliography{mybib}
}

\end{document}